\documentclass[conference]{IEEEtran}
\usepackage{cite}
\usepackage{amsmath,amssymb,amsfonts}
\usepackage{algorithmic}
\usepackage{graphicx}
\usepackage{textcomp}
\usepackage{xcolor}
\usepackage{todonotes}
\usepackage[hidelinks]{hyperref}
\usepackage{booktabs}
\def\BibTeX{{\rm B\kern-.05em{\sc i\kern-.025em b}\kern-.08em
    T\kern-.1667em\lower.7ex\hbox{E}\kern-.125emX}}
\usepackage{comment}
\usepackage[flushleft]{threeparttable}
\usepackage[autostyle]{csquotes}
\usepackage{listings}

\usepackage{fancyhdr}
\begin{document}

\title{Recognizing Emotion Regulation Strategies from Human Behavior with Large Language Models}

\author{\IEEEauthorblockN{Philipp M\"uller}
\IEEEauthorblockA{\textit{DFKI}\\
Saarbr\"ucken, Germany \\
philipp.mueller@dfki.de}
\and
\IEEEauthorblockN{Alexander Heimerl}
\IEEEauthorblockA{\textit{Augsburg University}\\
Augsburg, Germany \\
alexander.heimerl@uni-a.de}
\and
\IEEEauthorblockN{Sayed Muddashir Hossain}
\IEEEauthorblockA{\textit{DFKI}\\
Saarbr\"ucken, Germany \\
sayed\_muddashir.hossain@dfki.de}
\and
\IEEEauthorblockN{Lea Siegel}
\IEEEauthorblockA{\textit{DFKI}\\
Saarbr\"ucken, Germany \\
lea.siegel@dfki.de}
\and
\IEEEauthorblockN{Jan Alexandersson}
\IEEEauthorblockA{\textit{DFKI}\\
Saarbr\"ucken, Germany \\
jan.alexandersson@dfki.de}
\and
\IEEEauthorblockN{Patrick Gebhard}
\IEEEauthorblockA{\textit{DFKI}\\
Saarbr\"ucken, Germany \\
patrick.gebhard@dfki.de}
\and
\IEEEauthorblockN{Elisabeth Andr\'e}
\IEEEauthorblockA{\textit{Augsburg University}\\
Augsburg, Germany \\
elisabeth.andre@uni-a.de}
\and
\IEEEauthorblockN{Tanja Schneeberger}
\IEEEauthorblockA{\textit{DFKI}\\
Berlin, Germany \\
tanja.schneeberger@dfki.de}
}

\maketitle
\thispagestyle{fancy}

\begin{abstract}

Human emotions are often not expressed directly, but regulated according to internal processes and social display rules.
For affective computing systems, an understanding of how users regulate their emotions can be highly useful, for example to provide feedback in job interview training, or in psychotherapeutic scenarios.
However, at present no method to automatically classify different emotion regulation strategies in a cross-user scenario exists.
At the same time, recent studies showed that instruction-tuned Large Language Models (LLMs) can reach impressive performance across a variety of affect recognition tasks such as categorical emotion recognition or sentiment analysis.
While these results are promising, it remains unclear to what extent the representational power of LLMs can be utilized in the more subtle task of classifying users' internal emotion regulation strategy.
To close this gap, we make use of the recently introduced \textsc{Deep} corpus for modeling the social display of the emotion shame, where each point in time is annotated with one of seven different emotion regulation classes.
We fine-tune Llama2-7B as well as the recently introduced Gemma model using Low-rank Optimization on prompts generated from different sources of information on the \textsc{Deep} corpus.
These include verbal and nonverbal behavior, person factors, as well as the results of an in-depth interview after the interaction.
Our results show, that a fine-tuned Llama2-7B LLM is able to classify the utilized emotion regulation strategy with high accuracy (0.84) without needing access to data from post-interaction interviews.
This represents a significant improvement over previous approaches based on Bayesian Networks and highlights the importance of modeling verbal behavior in emotion regulation.

\end{abstract}

\begin{IEEEkeywords}
emotion regulation, large language models, emotion recognition, bayesian networks
\end{IEEEkeywords}

\section{Introduction}
\label{Introduction}

One key finding of emotion research is that there is no one-to-one mapping of displayed emotional expressions to internally experienced emotions~\cite{barrett2017emotions}. 
Emotions do not necessarily become visible~\cite{keltner1995signs}, nor consciously experienced~\cite{lewis2008self, gross2013handbook, nathanson1994shame}.
One reason for this is emotion regulation, which encompasses various conscious or unconscious strategies that individuals use to influence their emotional experience~\cite{gross1999emotion}. 
Especially unpleasant emotions such as shame are regulated to protect the self~\cite{lewis2008self, gross2013handbook, nathanson1994shame}.
For many affective computing systems, knowledge of users' emotion regulation strategies has the potential to be highly useful.
Such systems include social skill training systems~\cite{hoque2013mach, gebhard2018serious, schneeberger2021stress} or therapeutic assistance systems~\cite{gebhard2019designing, devault2014simsensei}. 
The recently introduced \textsc{Deep} approach was the first attempt to create a computational model of emotion regulation, focusing on the emotion shame elicited in job interviews~\cite{schneeberger2023DEEP}.
While the authors presented a Bayesian Network (BN) model to classify emotion regulation strategies, their approach had two key limitations prohibiting application in realistic scenarios. 
First, they require results of an extensive analysis of in-depth post-interaction interviews as input. 
Second, they did not evaluate their model in a subject-independent scenario.

Recent studies indicate that generative large language models (LLMs) are able to, in a certain sense, understand human emotion in social situations.
In zero-shot scenarios, GPT3.5 and GPT4 were successfully applied across a variety of emotion-related tasks, including sentiment analysis, emotion and emotion cause recognition, toxicity detection, and opinion extraction, albeit they are often still outperformed by approaches directly trained on the respective tasks~\cite{amin2023wide,zhao2023chatgpt}.
In contrast to zero-shot scenarios, instruction-tuning is an effective means to utilize the representational power of generative LLMs and at the same time adapt to a specific target task~\cite{li2024finetuning, zhang2024generation, borzunov2024distributed, liu2024understanding, zhang2023instruction}. 
Using Low-rank Adaptation (LoRA)~\cite{hu2021lora}, this process is computationally efficient, and was already utilized for tasks related to affect and social behavior~\cite{zhang2023dialoguellm,dey2024socialite,liu2024emollms}.
In particular, DialogueLLM~\cite{zhang2023dialoguellm} reached state-of-the-art results for emotion recognition on the MELD~\cite{poria-etal-2019-meld}, IEMOCAP~\cite{busso2008iemocap}, and EmoryNLP~\cite{zahiri2018emotion} datasets.
While these results are encouraging, it is unclear to what extent instruction-tuned LLMs can be used to classify emotion regulation strategies.
In contrast to expressions of emotion, these strategies reflect inner processes that may not have distinct observable cues and are believed to be heavily related to nonverbal aspects of behavior~\cite{schneeberger2023DEEP}.

In our work, we investigate to what extent instruction-tuned LLMs are capable of classifying the strategies employed by humans to regulate shame.
To this end, we make use of the recently introduced \textsc{Deep} corpus comprising recordings of human behavior in shame inducing situations and self-reported information about individual experience~\cite{schneeberger2023DEEP}.
Inspired by DialogueLLM~\cite{zhang2023dialoguellm}, we encode participants' multimodal behavior into prompts that are used for instruction-tuning Llama2-7B~\cite{touvron2023llama,meta-llama/Llama-2-7b-chat-hf} and Gemma~\cite{gemma_report_2024} models with LoRA~\cite{hu2021lora}.
We present the first cross-user evaluations on the \textsc{Deep} corpus and show that our LLM-based approach can reach an accuracy of 0.84 in emotion regulation classification without access to any information from informative but impractical post-interaction interviews.
As such, our results represent an important step towards affective computing systems that can recognize human emotion regulation strategies in realistic scenarios.

Our specific contributions are three-fold.
\begin{enumerate}
    \item We utilize LLMs instruction-tuned on prompts incorporating multimodal behavior to classify peoples' strategies to regulate the emotion shame.
    \item In the first cross-user evaluations on the recently introduced \textsc{Deep} corpus~\cite{schneeberger2023DEEP}, our approach outperforms the previous state of the art based on expert-constructed Bayesian Networks when information from post-interaction interviews is not available. %
    \item We conduct extensive ablation experiments, highlighting the impact of different modalities on performance.
\end{enumerate}

\section{Related Work}
\label{RelWork}

\subsection{Model of Emotions and Emotion Regulation}
\label{subsec:emotion_regulation}

There is a variety of emotion models both in psychology \cite{scherer2000psychological} and affective computing \cite{ps2017emotion}. %
In our work, we follow a model of emotions that differentiates between internal and external components inspired by cognitive psychoanalysis \cite{moser1996entwicklung}. 
\textit{Internal components of emotions} are not directly observable as they represent individual experience occurring in humans' inner worlds. Due to intrapersonal emotion regulation processes, the internal components may or may not be experienced consciously \cite{Tomkin84}. The intrapersonal emotion regulation, refers to how internal emotional components are managed  \cite{gross2013handbook}. It originates from psychoanalytical defense mechanism concepts and differs from the cognitive coping mechanism, which refers to a conscious-focused emotion regulation \cite{Cramer00}.
People regulate emotions to avoid or decrease experiential and/or behavioral aspects of negative emotions such as anger, sadness, and shame. Also positive emotions may be regulated -- for example, if the social situation requires it. 
The result of intrapersonal emotion regulation is the \textit{experienced component of emotions} and can be seen as the emotional information that is \enquote{bearable} within the related situation \cite{nathanson1992shame}.
\textit{External components of emotions} represent communicated information that regulates relationships with others and how they are experienced and represented internally. What is communicated externally is i.a. influenced by social display rules \cite{ekman2013repertoire}.
Due to both, intrapersonal emotion regulation and social display rules (interpersonal emotion regulation), 
the connection between internal and external components is not immediate and they do not necessarily match \cite{moser1996entwicklung, barrett2017emotions}.

For modeling human emotions computationally, computer scientists focused on cognitive appraisal theories for emotions \cite{Moorsetal13}. %
Some models take emotion regulation into account. 
One example is MARSSI \cite{gebhard2018marssi}, which models appraisal rules, emotion regulation rules, and social signal interpretation, and allows to define multiple possible and plausible relations between these components.
Furthermore, MARSSI differentiates between internal and external components of emotions.

Recently, \cite{schneeberger2023DEEP} presented the \textsc{Deep} method, a cognition-based method that focuses on modeling the internal component of emotions.
It incorporates an approach to query individual internal emotional experiences and to represent such information computationally. 
It combines social signals, with context information and information from a post-interaction interview (\enquote{verbalized introspection}).
These different components were modeled with a Bayesian Network constructed from theoretical domain knowledge.
They also presented first prediction results for the emotion regulation strategy employed by users.
However, their approach is limited in two key aspects which makes it impractical in many application scenarios.
First, it requires knowledge from the post-interaction interview, and second, it was not evaluated in a cross-subject scenario.
In contrast, we present instruction-tuned LLMs that are able to predict emotion regulation strategies with high accuracy in a cross-subject setting and without having access to information from the verbalized introspection collected post-interaction.

\subsection{LLMs and Emotion Understanding}
Large language models (LLMs) have been applied to a variety of tasks related to human affect expression~\cite{wang2023emotional,mao2022biases}.
One of the most popular of these tasks is sentiment analysis, which commonly involves classifying text into expressing a positive, negative, or neutral sentiment.
Transformer-based LLMs such as BERT, RoBERTa, or XLNet have been a key component of state-of-the-art sentiment analysis approaches in recent years~\cite{xie2020unsupervised,yang2019xlnet,wang2021entailment}.
With the success of generative LLMs such as GPT-3.5, GPT4, or Llama, researchers have investigated their utility for sentiment analysis, mainly in zero-shot and few-shot scenarios~\cite{mao2022biases,qin2023chatgpt}.
A slightly more complex task compared to sentiment analysis is categorical or dimensional emotion recognition.
Language models such as BERT or ROBERTa have been widely applied on these tasks~\cite{park2019dimensional,amin2023wide}.
GPT3.5 was shown to reach good performance on emotion- and emotion cause recognition, but is still outperformed by models fine-tuned for the specific task~\cite{zhao2023chatgpt}.
GPT4 improved upon GPT3.5 and is able to outperform an approach based on RoBERTa on tasks such as toxicity detection and opinion extraction, but it still lacks behind on tasks with strong implicit components such as subjectivity of personality estimation~\cite{amin2023wide}.
Emotion recognition in GPT-like models operating in zero-shot scenarios can be highly biased with respect to ground truth definition, prompt construction, or label word selection~\cite{mao2022biases}.

Recently, instruction tuning of large language models has become a popular technique to adapt generative LLMs to new tasks~\cite{zhang2023instruction}. 
By utilizing Low-rank Adaptation (LoRA)~\cite{hu2021lora}, fine-tuning models such as Llama2-7B became feasible on a single GPU.
This approach was also utilized for tasks related to affect and social behavior~\cite{zhang2023dialoguellm,dey2024socialite,liu2024emollms}.
In \cite{dey2024socialite}, authors used LoRA to create an instruction-tuned variant of Llama2-7B on various social behavior analysis tasks including, among others, sentiment and emotion classification.
In their experiments, instruction tuning leads to large performance gains relative to the standard Llama2 model.
In~\cite{liu2024emollms}, authors showed that instruction-tuned Llama2 models can clearly outperform all zero or few-shot approaches, including those based on GPT4 across a variety of affect recognition tasks.
The utilized emotion datasets of these approaches are entirely textual however, i.e. they do not incorporate nonverbal behavior present in a face-to-face interaction.
Despite the importance of nonverbal behavior for the expression of emotions, only few works have made attempts to include nonverbal behavior into the prompts given to LLMs~\cite{hasan2023textmi,zhang2023dialoguellm}.
In \cite{hasan2023textmi}, authors extracted textual descriptions from clusters of nonverbal behavioral features and used this information in addition to verbal input for sentiment analysis.
DialogueLLM \cite{zhang2023dialoguellm} classified emotions in conversation by constructing prompts describing the conversational and visual context, including nonverbal behavior of the interactants.
They fine-tuned Llama2-7B on several emotion recognition datasets, and outperform the previous state of the art on MELD~\cite{poria-etal-2019-meld}, IEMOCAP~\cite{busso2008iemocap}, and EmoryNLP~\cite{zahiri2018emotion}.
As such, instruction-tuning of LLMs seems to be a promising way to model interactions between verbal- and nonverbal behavior for emotion understanding tasks.
To the best of our knowledge, we for the first time apply instruction tuning on prompts generated from multimodal inputs to recognize emotion regulation strategies.

\begin{table}[t]
 \caption{Ground truth classes on the \textsc{Deep} Corpus \cite{schneeberger2023DEEP}, including their definition, as well as possible experienced components and nonverbal behaviour.}
\begin{threeparttable}
\begin{tabular}{p{0.45\textwidth}}
\toprule
       \textbf{WITHDRAWAL (655 frames)} Cut off the current situation so there is no more external influence or stimuli. Wish to hide, leave or escape.  \\
        Experienced emotional components: distress, fear \\ 
       {Nonverbal Behavior: freezing, lip biting, gaze/head aversion, silence} \\
         \midrule
        \textbf{ATTACK SELF (515 frames)} Do to yourself what others may do to you, establishing impression to control the situation. \\ 
       Experienced emotional components: disgust \\
       {Nonverbal Behavior: facial expression of disgust} \\
    \midrule
        \textbf{ATTACK OTHER (629 frames)} Transfer the diminishment of self-esteem to the person (object) who caused it by diminishing the other person. \\
       Experienced emotional components: anger \\
        {Nonverbal Behavior: learn forward, gestures of power, facial expression of anger}\\
    \midrule
        \textbf{AVOIDANCE (1650 frames)} Acting according the principle \enquote{fool others, fool myself}. \\
         Experienced emotional components: joy \\
        {Nonverbal Behavior: gaze/head aversion, lean backwards, facial expression of joy/surprise, smile} \\
    \midrule
       \textbf{DEPRECIATION (1911 frames)} Deevaluation of interaction partner due to different (or even contrary) values and ideals. \\ 
        Experienced emotional components: disgust, contempt \\
        {Nonverbal Behavior: raised eyebrows, smile, facial expression of disgust and contempt}\\
    \midrule
        \textbf{STABILIZE SELF (3593 frames)} Attempt to react in a way that is compliant with the (ideal) self by accepting disagreement between job interviewer and person. \\ 
         Experienced emotional components: pride \\
        {Nonverbal Behavior: no display of uncertainty, direct gaze}\\
    \midrule
         \textbf{REST (2582 frames)} No identified emotion regulation strategy.\\ 
    \bottomrule
\end{tabular}
    \label{tab:output}
\end{threeparttable}
\end{table}

\section{Corpus}

\begin{table*}[t]
    \centering
    \caption{Annotated input features on the \textsc{Deep}~\cite{schneeberger2023DEEP} corpus.}
    \begin{tabular}{p{3cm}p{14cm}}
    \toprule
       Nonverbal Behavior  & Observation of external components of emotions that are encoded in social signals in the specific situation. \\ 
       & \textit{Speech, Utterance, Facial Expression, Gaze, Eyes, Smile, Smile Control, Head, Head Tilt, Upper body, Shame display}  \\
       \midrule
       Verbalized introspection &  Self-reports that reflect a person's subjective experience gathered in semi-structured interviews after the specific situation with the aid of video material of the experienced situation.  \\
        & \textit{Relationship management, Shame awareness, Experienced emotion, Internal emotion component, Display rule}\\
         \midrule
       Personal Context  &  Personal context variables. \\ 
        & \textit{Gender, Mindedness score} \\
         \midrule
       Situational Context  &  Situational context variables. \\ 
        & \textit{Situation (first vs. second shame induction), Conversation transcript} \\
        \bottomrule
    \end{tabular}
    \label{tab:features}
\end{table*}

For our work, we utilize the recently introduced \textsc{Deep} corpus, which we received upon request from the authors~\cite{schneeberger2023DEEP}.
The corpus consists of shame-inducing situations during job interviews. 
It includes data from 20 expert-annotated videos of ten participants, each in two shame-eliciting situations, comprising 11535 video frames.
Shame was elicited in mock job interviews framed as job interview trainings. During these, participants were confronted with a virtual job interviewer (avatar). To elicit shame in participants, the following validated, controlled and pre-evaluated situations~\cite{schneeberger2019can} were:
\begin{enumerate}
    \item After greeting the interviewee, the job interviewer says: \enquote{Before we start, a quick question. Where did you get that outfit? Somehow it doesn't really suit you.} Following \cite{nathanson1994shame}, this statement reflects the association \textit{personal attractiveness} to the self.
     \item After the interviewee has presented their experience, the interviewer reacts as follows: \enquote{All the other applicants have already said what you said. You haven’t exactly stood out.} Following \cite{nathanson1994shame}, this statement reflects the association \textit{Sense of self}.
\end{enumerate}

After the interaction with the avatar, participants went through an interview reflecting about their experience, called the ``verbalized introspection''.
The \textsc{Deep} corpus consists of data from different sources of information about each specific shame-eliciting situation. 
Annotations were done by three trained raters (all with a degree in psychology, one of them an experienced psychotherapist) based on the behavior of the participant in the shame-eliciting interview, the transcribed verbalized introspection, the context and the theoretical knowledge about shame and shame regulation. 
We utilize the annotations from \cite{schneeberger2023DEEP} as ground truth as well as input features.
As ground truth, we use the seven emotion emotion regulation strategy classes (see \autoref{tab:output} for an overview). 
In the \textsc{Deep} corpus, primary and secondary emotion regulation strategy annotations exist, as -- similar to emotions \cite{harris1985children} -- several emotion regulation processes can be active at the same time. 
For the purpose of this paper, we chose to focus on the primary emotion regulation strategy exclusively.
The input features consist of annotations extracted from nonverbal behavior, verbalized introspection, personal context, and situational context (\autoref{tab:features}).
For the purpose of this paper, we transcribed participants' verbal answers in the shame eliciting situations and added these to the situational context features. 
For further information on the corpus and the different annotations, we refer to the Supplemental Material of~\cite{schneeberger2023DEEP}.

 \begin{figure}[t]
 \centering
  \includegraphics[width=0.47\textwidth]{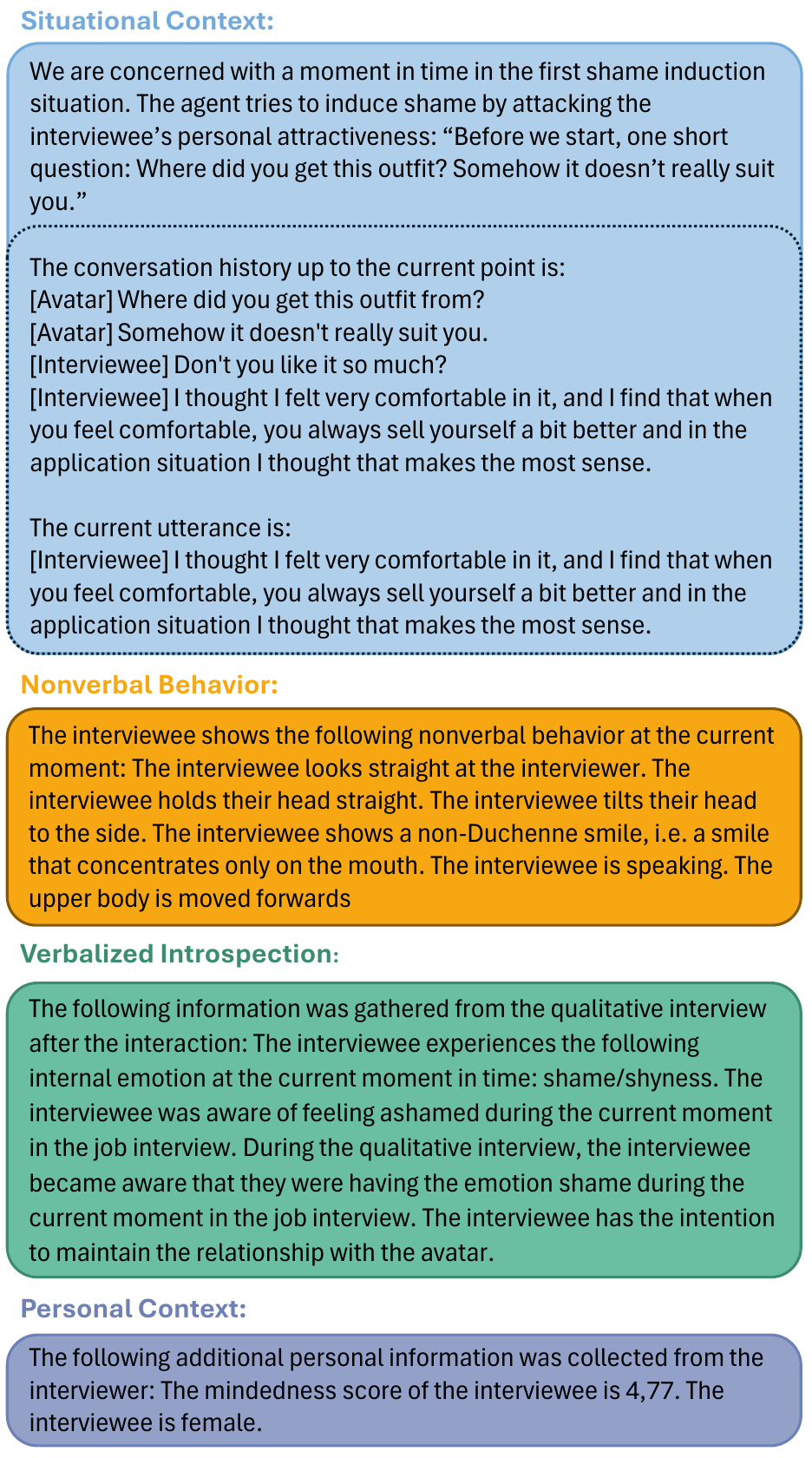}
  \caption{Example prompt consisting of situational context, nonverbal behavior, verbalized introspection and personal context. The situational context incorporates a transcript (below the dotted line).}
  \label{fig:prompt}
\end{figure}

\section{Approach}
\label{Approach}

We preset our approach based on instruction-tuned Large Language Models (LLMs), as well as our baseline implementation of the Bayesian Networks proposed in~\cite{schneeberger2023DEEP}.

\subsection{Multi-modal LLM Approach}

Our approach uses instruction tuning to fine tune LLMs on prompts created from different sources of information, including verbal and nonverbal behavior as well as contextual information.

\subsubsection{Prompt generation}
We construct one prompt from every frame in the corpus.
Similar to~\cite{zhang2023dialoguellm}, we generate textual descriptions from different sources of information.
An example prompt, broken down into components, is shown in \autoref{fig:prompt}.
In particular, we provide situational context by describing the particular shame induction situation, and providing a transcript of the verbal exchange up until the current frame.
We also clearly define the utterance for which the model is supposed to classify the shame regulation strategy, i.e. the utterance corresponding to the current frame.
The nonverbal behavior annotated on the Deep corpus at the current frame is directly translated to textual descriptions. 
E.g. annotation ``TILT'' for interviewee head behavior is annotated, that would translate to ``The interviewee tilts their head to the side''.
Finally, we add a textualization of the personal context variables.
The results of verbalized introspection are not part of our default approach, however as we add them in certain experiments, they are included for reference in \autoref{fig:prompt}.
For the verbal prompt components, we translated the German transcripts on the \textsc{Deep} corpus to English, using the mbart-large-50-many-to-many-mmt model \cite{tang2020multilingual}. 
This model's multilingual capabilities enabled it to surpass the performance of several one-to-one translation models. 
Our experiments confirmed this; we initially tested the smaller opus-mt-en-de model \cite{TiedemannThottingal:EAMT2020}, but its translations were notably inferior to those produced by the mbart-based model after careful review.

\subsubsection{Context information}
In addition to the prompt generated from each frame, we provide constant context information to the model, explaining the task, situation, and ground truth definitions (i.e. extended definitions of the shame regulation strategies shown in~\autoref{tab:output}).
We include this context information in the supplementary material.

\subsubsection{Utilized LLMs}
We utilized a variant of the Llama LLM \cite{touvron2023llama} specifically, the Llama-2-7b-chat-hf model \cite{meta-llama/Llama-2-7b-chat-hf}. We opted for this chat-oriented model as its fine-tuning on conversational data enhances its ability to understand the nuances of human dialogue. Preliminary experiments comparing the base Llama model and the chat variant supported this decision, with the latter yielding superior results.
Additionally, we incorporated the recent Gemma model~\cite{gemma_report_2024} from Google DeepMind, which reached state-of-the-art performance across various NLP tasks.

\subsubsection{Training Details}
For training, the inputs were the Prompt and the Context. 
The relevant output was the emotion regulation strategy.
To fine-tune our models, we applied the Low-Rank Adaptation of Large Language Models (LoRA) technique \cite{hu2021lora}.  
For LoRA, we follow previous work~\cite{dettmers2024qlora} and set $r=8$, $\alpha=16$ and dropout of 0.1.
Further hyperparameters are documented in the supplementary material.
We trained both models for 5 epochs.
For Llama2-7B we were able to use 16 batches per device, for Gemma only 4. 
Further training details and code are available online\footnote{\url{https://git.opendfki.de/philipp.mueller/acii24_emotionregulationllm}}.
In total, we made use of three Nvidia A100 GPUs with 40GB VRam each: two for fine-tuning and one for test-time inference.
Training lasted for about 2 weeks to generate all results in this paper.

\subsubsection{Testing}
For testing, the model was put first into inference mode, with zero temperature. 
Then, the Prompt and Context were were fed to the model for each of the instances. 
We extracted the predicted emotion regulation strategy from the model's response. 
We checked for anomalies in the response of the LLMs via string matching between the generated and the set of desired output classes, but both LLMs always predicted exactly one emotion regulation strategy label for each sample. 

\subsection{Bayesian Network Model}
\label{BN_model}
\begin{figure}[t]
\centering
  \includegraphics[width=.490\textwidth]{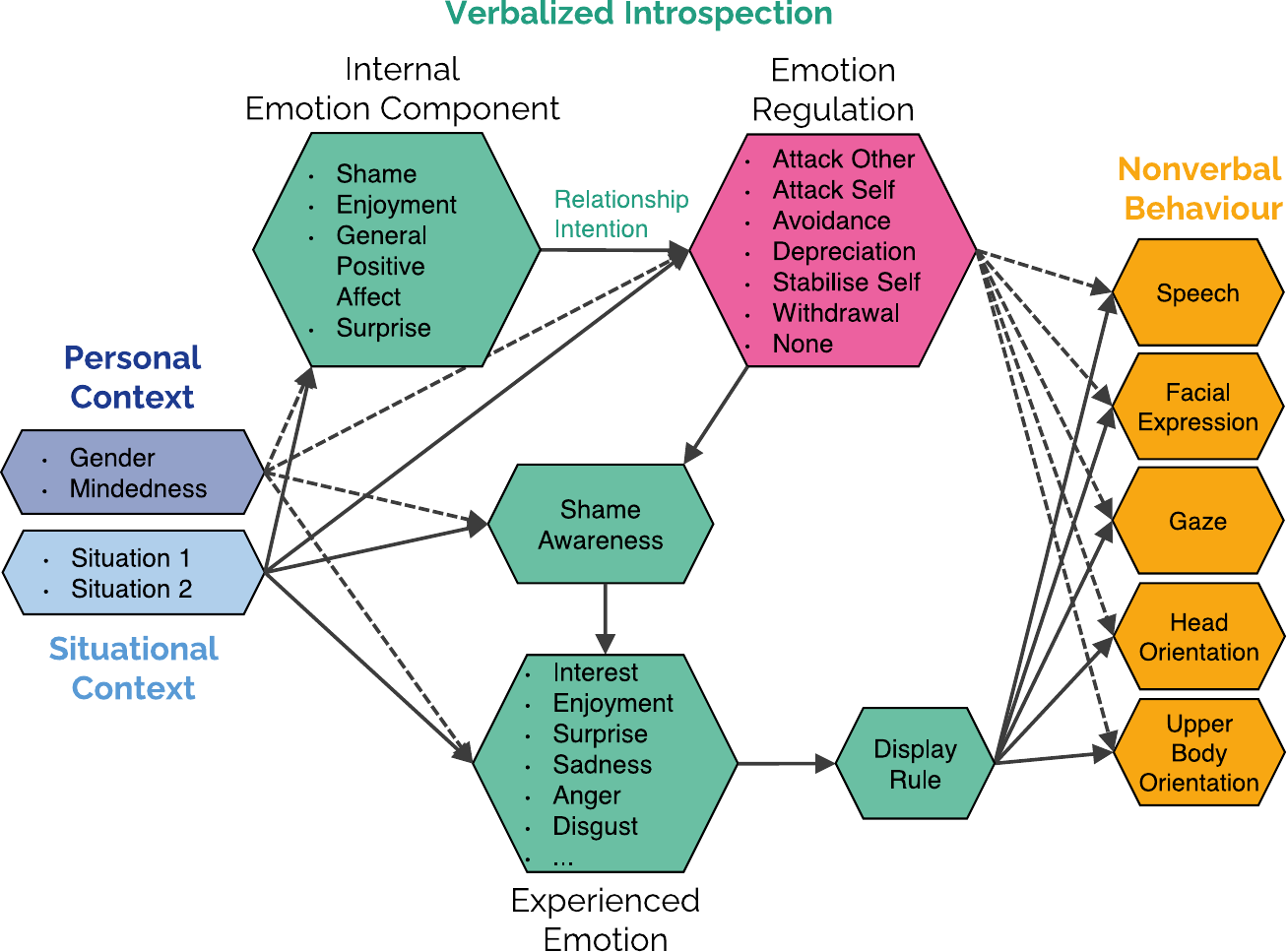}
  \caption{The \textsc{Deep}-BN schema constructed based on the \textsc{Deep} method information. The ground truth of emotion regulation is also part of the verbalized introspection. But since it represents the ground truth (and not a potential input to the model), it is colored pink. The emotion regulation strategies are based on \cite{nathanson1994shame}, while internal emotion component and the experienced emotion are based on the Differential Emotions Scale \cite{izard1974differential} and PANAS-X \cite{watson1994panas}.}
  \label{fig:deepdbn}
\end{figure}

As a baseline comparison, we make use of the \textsc{Deep}-BN approach proposed in \cite{schneeberger2023DEEP}.
This method conceptualizes a Bayesian network (BN) model representing the Internal Emotion Component, Emotion Regulation and related concepts (see \autoref{fig:deepdbn}). 
Bayesian Networks are graphical models and, compared to other Machine Learning frameworks relatively easy to comprehend and therefore are ideal for modeling theory-based implications and their explanation.

In general, there are two types of nodes in the \textsc{Deep}-BN. Blue nodes in the Figure represent information that is updated based on observations in the BN, red nodes represent information that is inferred by the BN. When it comes to understanding the internal emotions it is essential to model the interplay between the Internal Emotion Component, the process of Emotion Regulation, the Experienced Emotion Component, and related Social Signals. 

The Internal Emotion Component represents possible emotion classes that -- depending on the Emotion Regulation -- may or may not result in a consciously Experienced Emotion Component. It is possible that individuals do not apply strong Emotion Regulation which results in a match between the Internal Emotion Component and the Experienced Emotion Component. However, it may also be that the Emotion Regulation is strong and unconscious resulting in a completely different Experienced Emotion Component compared to the Internal Emotion Component (e.g., Experiencing anger when unconsciously applying the Emotion Regulation strategy Attack Other but not shame) (see Sec.~\ref{subsec:emotion_regulation}).
The Social Signals represent the observable result of the underlying Experienced Emotion Component and applied Emotion Regulation.

The Internal Emotion Component, the Emotion Regulation and the Experienced Emotion Component are influenced by the Personal Context, for example, demographic aspects (e.g., gender), or personality aspects (e.g., mindedness), as well as the Situational Context (i.e. the shame-inducing situation). 

The BN we built based on the \textsc{Deep} method acts as a benchmark to investigate the capabilities of LLMs to recognize emotion regulation strategies. Even though the main focus of the \textsc{Deep} method is to provide a deeper understanding of the Internal Emotion Component, the architecture of a BN allows us to easily change the inference target from predicting internal emotions to predicting emotion regulation strategies given a specific emotion.

\begin{table*}[]
\caption{Accuracy and F1-Score of different models for emotion regulation recognition on different ground truth classes, as well as overall. As overall F1 score, we report the weighted F1 score.}
\begin{tabular}{l|cccccccccccccccc}
& \multicolumn{2}{c}{Withdrawal} & \multicolumn{2}{c}{Attack self} & \multicolumn{2}{c}{Attack other} & \multicolumn{2}{c}{Avoidance} & \multicolumn{2}{c}{Depreciation} & \multicolumn{2}{c}{Stabilize self} & \multicolumn{2}{c}{Rest} & \multicolumn{2}{c}{Overall}\\ 
\cmidrule(lr){2-3} \cmidrule(lr){4-5} \cmidrule(lr){6-7} \cmidrule(lr){8-9} \cmidrule(lr){10-11} \cmidrule(lr){12-13} \cmidrule(lr){14-15} \cmidrule(lr){16-17}
& ACC & F1 & ACC & F1 & ACC & F1 & ACC & F1 & ACC & F1 & ACC & F1 & ACC & F1 & ACC & F1 \\ 
\midrule
\textit{w/ verb. introspection} \\
\ \ \ \ Bayesian Net & \textbf{0.99} & \textbf{0.94} & \textbf{0.99} & \textbf{0.91} & \textbf{0.99} & \textbf{0.93} & \textbf{0.99} & \textbf{0.99} & \textbf{0.98} & \textbf{0.94} & \textbf{0.98} & \textbf{0.98} & 0.96 & 0.91 & \textbf{0.96} & \textbf{0.96} \\
\ \  \ \ Gemma & 0.98 & 0.83 & \textbf{0.99} & 0.86 & 0.98 & 0.86 & 0.98 & 0.92 & \textbf{0.98} & 0.93 & 0.97 & 0.95 & \textbf{0.98} & \textbf{0.95} & 0.93 & 0.93\\
\ \  \ \ Llama2-7B & \textbf{0.99} & 0.88 & 0.97 & 0.69 & 0.98 & 0.84 & 0.98 & 0.94 & 0.96 & 0.89 & 0.95 & 0.92 & 0.95 & 0.88 & 0.89 & 0.89\\
\midrule
\textit{w/o verb. introspection} \\
\ \ \ \ Bayesian Net & 0.81 & 0.21 & 0.88 & 0.0 & 0.89 & 0.08 &  0.79 & 0.33 & 0.65 & 0.13 & 0.69 & 0.34 & 0.72 & 0.26 & 0.23 & 0.25\\
\ \  \ \ Gemma & 0.94 & 0.56 & 0.94 & 0.55 & 0.94 & 0.57 & 0.92 & 0.70 & 0.90 & 0.70 & 0.88 & 0.78 & 0.90 & 0.76 & 0.71 & 0.72\\
\ \  \ \ Llama2-7B & \textbf{0.97} & \textbf{0.76} & \textbf{0.97} & \textbf{0.71} & \textbf{0.96} & \textbf{0.71} & \textbf{0.96} & \textbf{0.85} & \textbf{0.95} & \textbf{0.84} & \textbf{0.93} & \textbf{0.88} & \textbf{0.95} & \textbf{0.88} & \textbf{0.84} & \textbf{0.84}\\
\end{tabular}
\centering
\label{tab:results}
\end{table*}

\begin{table*}[]
\caption{Weighted Accuracy and weighted F1-Score of Bayesian Networks for emotion regulation recognition containing different modalities}
\begin{tabular}{l|cccccc}
 & \multicolumn{2}{c}{Bayesian Net} &  \multicolumn{2}{c}{Llama2-7B} & \multicolumn{2}{c}{Gemma} \\ 
 \cmidrule(lr){2-3} \cmidrule(lr){4-5} \cmidrule(lr){6-7}
 Input Modalities & ACC & F1 & ACC & F1 & ACC & F1 \\ 
 \midrule
\textit{w/ verb. introspection} \\
\ \ \ \ All & \textbf{0.96} & \textbf{0.96} & 0.89 & 0.86 & 0.93 & 0.93\\
\ \ \ \ No personal context & 0.69 & 0.68 & 0.88 & 0.88 & 0.93 & 0.93\\
\ \ \ \ No situational context & 0.84 & 0.85 & 0.49 & 0.51 & 0.61 & 0.63\\ 
\ \ \ \ No transcript & --- & --- & 0.45 & 0.47 & 0.63 & 0.64\\
\ \ \ \ No nonverbal behavior & 0.06 & 0.01 & 0.87 & 0.87 & 0.87 & 0.87\\
\ \ \ \ Only verbalized introspection & 0.17 & 0.16 & 0.54 & 0.56 & 0.54 & 0.56\\
\midrule
\textit{w/o verb. introspection} \\
\ \ \ \ All & 0.23 & 0.25 & \textbf{0.84} & \textbf{0.84} & 0.71 & 0.72\\
\ \ \ \ No personal context & 0.26 & 0.27 & 0.44 & 0.46 & 0.45 & 0.47 \\
\ \ \ \ No situational context & 0.22 & 0.23 & 0.38 & 0.40 & 0.35 & 0.38 \\
\ \ \ \ No transcript & --- & --- & 0.40 & 0.42 & 0.34 & 0.37 \\
\ \ \ \ No nonverbal behavior & 0.25 & 0.28 & 0.42 & 0.44 & 0.44 & 0.46 \\
\ \ \ \ Only nonverbal behavior & 0.25 & 0.25 & 0.47 & 0.50 & 0.44 & 0.46\\
\end{tabular}
\centering
\label{tab:f1_acc_modalities}
\end{table*}

\section{Evaluation}
\label{Evaluation}
We trained and evaluated the approaches discussed in \autoref{Approach} on the task of classifying the user's emotion regulation strategy for each frame during the shame induction situations on the \textsc{Deep} corpus~\cite{schneeberger2023DEEP}.
To assess the generalizability of the %
models we employed a LOSO (leave-one-subject-out) evaluation. 
We evaluate our models in two general settings: (1) with verbalized introspection, i.e. including the information gathered from the post-interaction interview, and (2) without verbalized introspection.
While we expect the first setting to reach higher performance, the second setting respects the demands of application scenarios, where it is usually impractical to perform an additional interview with the user.

\subsection{Overall Results}
\autoref{tab:results} presents the models' accuracy and F1 score for each class for our two evaluation scenarios, i.e. with verbalized introspection, and without verbalized introspection. 
When considering all available information including the verbalized introspection the three models achieved excellent accuracy and F1 scores. 
However, the BN slightly outperformed the two LLMs in terms of overall accuracy and F1 score, with 0.96 and 0.96 respectively. 
The BN achieved the highest accuracy and F1 scores for all classes except the Rest class, here the Gemma model was able to surpass the BN with an accuracy of 0.98 and F1 score of 0.95. 
However, when excluding the information about the verbalized introspection the predictive performance of the BN heavily decreased. 
The BN was only able to achieve an overall accuracy of 0.23 and F1 score of 0.25. 
In contrast to that, the LLMs were still able to largely maintain their performance. 
The Llama2-7B model outperformed the Gemma model for both metrics with an accuracy of 0.84 and a f1-score of 0.84 in comparison to an accuracy of 0.71 and F1 score of 0.72.
In addition to comparing the predictive performance of the three models when including or excluding the information about the verbalized introspection we also investigated the influence of the other modalities on the recognition scores. 
When inspecting the per-class F1 scores we observe a slight trend towards lower performances for less frequent classes across all models.
In the case of Llama2-7B without verbalized introspection, F1 scores for Withdrawal (655 frames), Attack self (515 frames), and Attack other (629 frames) are between 0.71 and 0.76, whereas for the remaining classes (each $>1500$ frames) they range from 0.84 to 0.88.

In preliminary experiments, we investigated the feasibility of a zero-shot approach without instruction tuning based on Llama2-7B.
We made two observations.
First, we were not able to instruct the model to output a classification decision instead of a text generation, making this approach impractical for full-scale quantitative evaluations.
Second, on a small test set of five samples from each ground truth class, we observed that the model's outputs are highly biased: in 30 out of 35 cases the model predicted Stabilize self.

\subsection{Ablation Results}
\autoref{tab:f1_acc_modalities} displays the overall accuracy and weighted F1 score for the three classifiers considering different modalities. 
When considering verbalized introspection and all other available modalities we already reported that the BN performed the best with the highest scores overall. However, when removing information about nonverbal behavior or even only considering the verbalized introspection the recognition scores of the BN drastically decrease. 
The removal of nonverbal behavior has very little influence on the accuracy and F1 score of both LLMs. 
But removing situational context (which includes the transcript) leads to a noticeable decrease in prediction performance for the Llama2-7B and Gemma models. 
In fact, the accuracy and F1 score similarly decrease as when excluding the transcript only (but keeping the information about the shame inducing situation), indicating that the key information the LLMs utilize is users' verbal behavior.
For the BN, the information about the situational context is less important to correctly predict the emotion regulation strategies. 
Without access to verbalized introspection, the BN only reaches F1 scores between 0.23 to 0.28, while the LLMs can better maintain their performance.
As in the condition with available verbalized introspection, removal of situational context or transcript impacts the LLMs most.

\section{Discussion}

\label{Discussion}

\subsection{On Performance}
While the Bayesian Network based approach achieved the highest performance when all modalities including verbalized introspection were available, the LLMs where much more robust when modalities were removed.
Especially the fact that LLMs proved to be relatively robust to the removal of verbalized introspection information makes them a decidedly better choice in application scenarios where post-interaction interviews are impractical, or online prediction is desired.

It is crucial to acknowledge that the Bayesian Network (BN) does not include the raw transcript of the job interview, but a distilled representation of the data sourced from the job interview and verbalized introspection.
The distillation can be beneficial, especially if it encapsulates the most pertinent information required to identify affective states. 
However, there's a risk that during this process, potentially relevant details may be excluded. 
Thus, depending on the quality of abstraction, the removal of modalities may have a less or more detrimental impact on the performance of the BN. 
For example, the internal emotion component appears to contain the most relevant information by representing the extracted emotion classes. 
In our case, leaving out the information associated with that component has a detrimental impact on the performance of the BN which cannot be compensated by the information associated with the situational context. 
This underscores the critical nature of the abstraction approach, particularly in how omitted information impacts the comparative efficacy of the BN and LMM in emotion recognition tasks.

LLMs bear the advantage that they are able to access semantic information from the transcripts of the job interviews. 
This information enables them to leverage additional nuanced information crucial for affect recognition while the BN has only access to this information in terms of abstract representations gathered from the verbalized introspection. 
While the BN benefits from incorporating a theory-driven emotion model, the advantage of such a model is contingent upon its access to relevant information resulting from verbal introspection.

\subsection{Limitations and Future Work}
While our results represent an encouraging step towards emotion regulation recognition in realistic scenarios, several limitations remain. 
Due to the need for verbalized introspection and the complexity of the annotations, the \textsc{Deep} corpus is limited in size and variability.
The ten participants were all having the same cultural background, similar age and were pre-selected having good skills to reflect on their internal experiences. 
Therefore, the full range of emotion regulation strategies and associated nonverbal behavior may not be captured, which may limit the generalizability of our findings.
The reduction of effort by using an LLM to predict emotion regulation strategies, where verbalized introspection seems to be less crucial, seems promising. 
It would allow for more economical data collection and annotation for future work investigating emotion regulation strategies, however the accuracy of the LLM predictions need to be rigorously evaluated in any new scenario.

This paper focuses on emotion regulation in validated shame-eliciting situations, limiting the extension of the work to situations where other emotion classes are elicited. 
Shame is an ideal starting point for this kind of research, both because of the existing extensive theoretical background describing shame regulation strategies~\cite{nathanson1994shame}, as well as due to the availability of the \textsc{Deep} corpus. 
However, emotions are not only regulated in shame eliciting situations, as most (if not all) emotions are intrapersonally regulated \cite{Tomkin84}.
Therefore, future work should extend the application of this proposed hybrid approach to other emotion classes, to gain an overall deeper understanding of individual emotional experiences. 

Finally, while our proposed approach allows to automatically infer emotion regulation strategies from behavioral descriptions, the descriptions provided with the \textsc{Deep} dataset were manually annotated.
Future work should replace such manual steps with automatic methods.
While this might not be easy to do for features extracted from the verbalized introspection, automatic methods to detect facial behavior~\cite{openface2}, body language~\cite{balazia2022bodily}, and to recognize speech~\cite{radford2023robust} are available.
When using automatic approaches, the set of nonverbal behaviors can also easily be extended, e.g. by detecting backchannels~\cite{amer2023backchannel}, or analyzing prosody~\cite{eyben2015geneva}.

\section{Conclusion}
In this paper, we presented the first evaluation of instruction-tuned large language models (LLMs) on the task of recognizing the strategy employed to regulate the emotion shame.
We utilized the recently introduced \textsc{Deep} corpus of shame-inducing situations during job interviews, which is annotated with multi-modal behaviors and verbalized introspection gathered after the shame-inducing interactions.
Our results indicate that while theory-driven Bayesian Networks perform best when all information is available, LLMs can cope much better with missing information from the verbalized introspection, likely due to their capability to effectively make use of users' verbal behavior.
As such, our insights are an important building block towards affective computing systems able to recognize emotion regulation strategies in realistic scenarios.

\section*{Ethical Impact Statement}

The paper employs data from the recently introduced DEEP corpus, which we received upon request from the authors. The DEEP corpus includes recordings of human behaviors in job interviews and subsequent verbal introspection. The collection and analysis of such data involves processing personal and potentially sensitive data. Furthermore, it exposes participants to shameful situations which may lead to negative emotional states. It is crucial to obtain informed consent from the participants, ensure that the employed stimuli don't negatively affect their mental health, implement robust data protection measures and only collect data necessary for the intended affect recognition purposes. Approval for collecting and processing these data was obtained from the ethical review board of the DEEP corpus’ authors. The current analysis of multimodal interview data and subsequent verbal introspection is covered by the ethics’ approval for the DEEP corpus.

Our model contributes to endeavors aimed at deciphering internal states, particularly benefiting Affective Computing systems reliant on discerning user emotions, such as social training systems or therapeutical assistants. However, the potential for misapplication raises pertinent privacy concerns.

In our investigation, we solicited insights from participants concerning their internal experience in shame-eliciting situations. Although participants provided consent for research purposes, in practical scenarios, individuals may withhold consent due to apprehensions surrounding the exposure of their internal experiences. Such reluctance could engender adverse ramifications for social interactions and interpersonal relationships.

Prior to engaging with systems employing models for interpreting observable expressions and internal states, it is imperative that users are adequately informed and provide consent regarding functionality, data collection, processing, and attendant risks. The utilization of such systems without the informed consent of individuals subject to observation may result in deleterious outcomes. Unsanctioned application of such technologies may inadvertently gather deeply personal information about individuals' internal experiences, subsequently exposing them to potential harm to their social standing, privacy, and overall well-being.

\section*{Acknowledgment}
P. M\"uller, S. Hossain, L. Siegel, and J. Alexandersson were partially funded by  the European Union Horizon Europe programme, grant number 101078950. 
E. Andr\'e, P. Gebhard, and T. Schneeberger were supported by the German Federal Ministry for Education and Research (BMBF) as a segment of the UBIDENZ project,
under grant numbers 13GW0568D and 13GW0568F.

\bibliographystyle{ieeetr}
\bibliography{bibliography}

\end{document}